\documentclass[10pt,twocolumn,letterpaper]{article}

\usepackage{cvpr}      
\usepackage{arydshln}
\usepackage{multirow}
\usepackage{tabularx}
\usepackage{array}

\usepackage{amsmath,amssymb,graphicx}
\usepackage{amsfonts}
\usepackage{url}
\usepackage{algorithm}
\usepackage{algorithmicx}
\usepackage{algpseudocode}
\usepackage{booktabs}
\usepackage{adjustbox}
\usepackage{threeparttable}
\usepackage{multirow}
\usepackage[table]{xcolor}%

\usepackage{array}
\usepackage{amsthm,amsmath,amssymb}
\usepackage{mathrsfs}

\usepackage{psfrag}
\usepackage{epsfig}
\usepackage{rotating}
\usepackage{colortbl}
\usepackage[table]{xcolor} 

\usepackage{diagbox}

\usepackage{siunitx}  

\usepackage[accsupp]{axessibility}  

\usepackage{url}

\definecolor{deepgreen}{RGB}{0,100,0} %

\usepackage{bbm}
\usepackage{wrapfig}
\usepackage{bbding}
\usepackage{amssymb} 
\usepackage{comment}
\usepackage{makecell}

\usepackage{colortbl}

\definecolor{cvprblue}{rgb}{0.21,0.49,0.74}
\usepackage[pagebackref,breaklinks,colorlinks,allcolors=cvprblue]{hyperref}

\def\paperID{7670} 
\def\confName{CVPR}
\def\confYear{2026}

\title{Reversing the Flow: Generation-to-Understanding Synergy in Large Multimodal Models}

\author{
    Yujun Tong$^{1,2}$~~
    Dongliang Chang$^{1,2}$\thanks{Corresponding author.}~~ 
    Zijin Yin$^{1,2}$~~ 
    Xintong Liu$^{1,2}$~~  
    Yuanchen Fang$^{1,2}$~~ 
    Zhanyu Ma$^{1,2}$ \\
    {\small $^{1}$Beijing University of Posts and Telecommunications} \\
    {\small $^{2}$Beijing Key Laboratory of Multimodal Data Intelligent Perception and Governance} \\
    {\tt\small \{tongyujun, changdongliang, yinzijin2017, liuxintong, fangyuanchen, mazhanyu\}@bupt.edu.cn}
    }
\begin{document}
\maketitle

\begin{abstract}

The long-standing goal of multimodal AI is to build unified models in which visual understanding and visual generation mutually enhance one another. Despite recent works such as BAGEL, BLIP3o achieves remarkable progress; In practice, however, this unification remains one-directional: understanding routinely guides generation, yet how and why generation can support understanding is rarely investigated.
We revisit this asymmetry and propose \textbf{Generation-to-Understanding (G$\rightarrow$U) synergy}, where visual generation becomes an explicit intermediate reasoning step.  
Our framework enables a model to perform controlled generative acts, such as detail enhancement, context expansion or structural visualisation, to produce self-generated \textbf{visual thoughts}, which are then fed back into the model to refine perception without retraining or external tools.
Through a comprehensive evaluation on twelve benchmarks, this reversed information flow consistently improves multimodal understanding.  
We show that generative fidelity bounds perceptual gain and that distinct families of edit prompts govern transfer efficiency.  
We further analyse whether models can decide what to imagine. While they can produce plausible edits, these self-generated visual thoughts lack stable task alignment, revealing that current large multimodal models fall short of true self-reflection.
This work exposes a missing mechanism in unified cognition and suggests that imagination is not the end of understanding but its beginning.

\end{abstract}

\section{Introduction}
\label{sec:intro}

Large multimodal models (LMMs) have made striking progress in recent years.  
Models such as GPT-5~\cite{openai2025gpt5}, Gemini~\cite{gemini}, and QwenVL~\cite{qwen25vl} now demonstrate strong visual reasoning, while diffusion models like SDXL~\cite{podell2023sdxl} and DALL$\cdot$E~3~\cite{openai2023dalle3} deliver unprecedented fidelity and controllability in image synthesis.  
The long-standing goal of multimodal AI is to unify these abilities within a single model that can both perceive and create.  
Such a model would reason through perception and verify through generation, closing the loop between understanding and synthesis.

\begin{figure}[!t]
\begin{center}
   \includegraphics[width=1\linewidth]{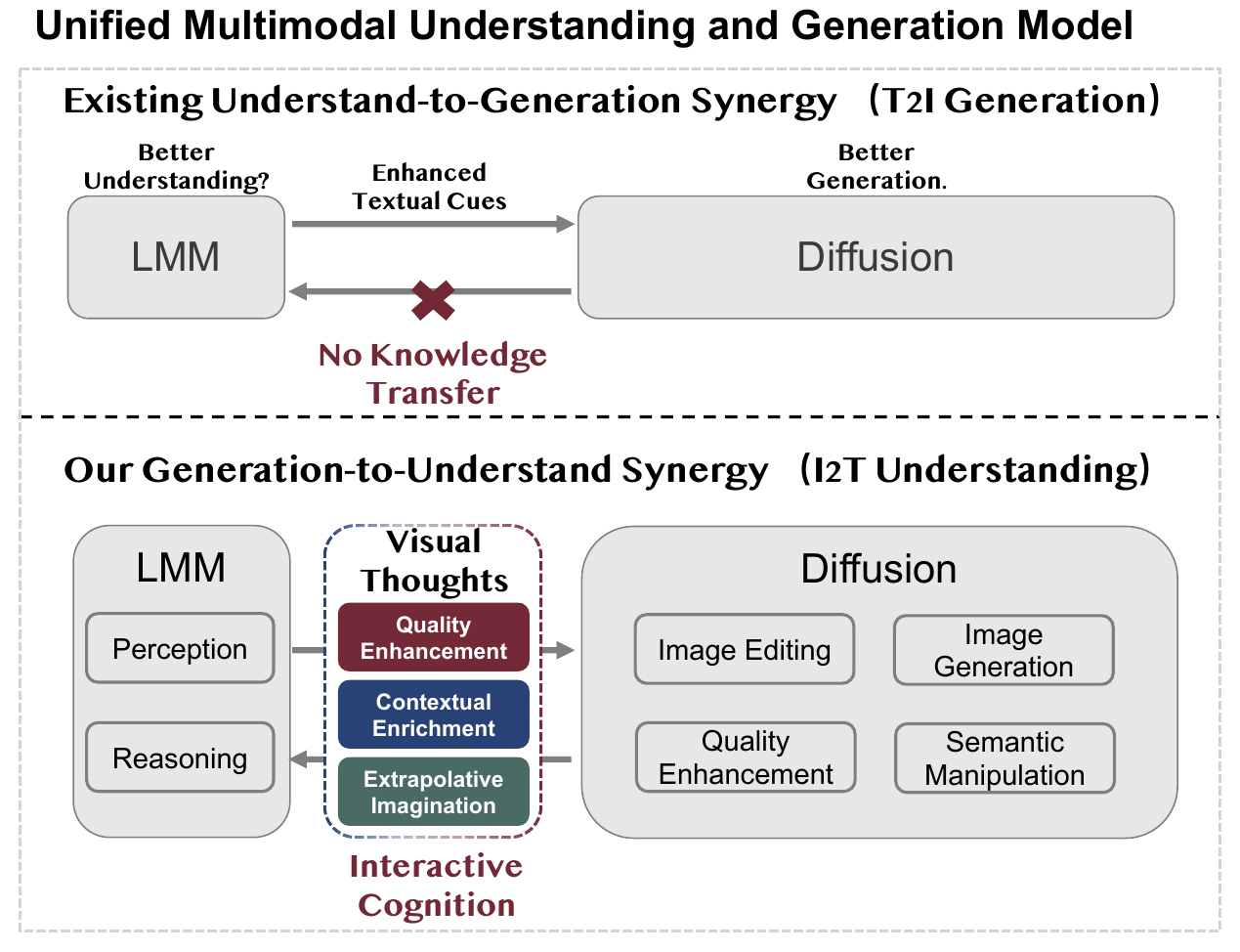}

\caption{\textbf{The core ambition of unified models is a true synergy where understanding (U) and generation (G) mutually reinforce.} However, current unification is predominantly one-directional (U $\rightarrow$ G), as an LMM's (e.g., QwenVL) strong reasoning is leveraged to high-fidelity generation. The reciprocal path (G $\rightarrow$ U), where generative faculties enhance understanding, remains critically overlooked. This work explores this missing link, demonstrating how a model's intrinsic generative capabilities can ``imagine" visual clues to bolster its own comprehension.} \label{fig:insight}
\end{center}
\vspace{-5mm}
\end{figure}

Current unified architectures appear to move toward this goal.  
They integrate autoregressive reasoning and diffusion-based generation within shared parameters~\cite{deng2025emerging,shi2024lmfusion,chen2025blip3o}.  
They can both understand and generate, often within a single transformer.  
Yet these abilities remain parallel rather than cooperative.  
Reasoning guides generation, but generation never feeds back into reasoning.  
The unification is structural rather than cognitive.

This asymmetry stems from the pipeline followed by all existing systems.  
A vision or vision–language backbone first interprets the input, producing semantic features.  
These features then condition a diffusion or autoregressive decoder that produces an output image~\cite{pan2025metaquery}.  
This design defines the prevailing understanding-to-generation paradigm that supports models such as InstructPix2Pix~\cite{brooks2023instructpix2pix}, Show-o~\cite{xie2024show}, BAGEL~\cite{deng2025emerging}, and Janus~\cite{wu2025janus,ma2025janusflow,chen2025januspro}.  
It enables impressive visual editing, but the reverse direction where generative faculties enhance perception has not been realised (See Figure~\ref{fig:insight}) . 
Moreover, continual training on generation-oriented data often strengthens synthesis at the cost of weakening the model’s original understanding skills.  
Thus, even when reasoning and synthesis share parameters, their interaction remains fundamentally one-way.  
The field has spent years teaching models to generate from understanding, but has seldom asked whether generation itself could teach understanding.

Humans, however, do not treat imagination as an output.  
When perception is uncertain, we imagine that we understand.  
We reconstruct missing details, visualise alternative viewpoints, and simulate context until meaning becomes clear.  
This raises a simple but largely unexplored question:  
\textit{Can a model use its own generative capability to improve its understanding?}

We explore this missing direction and introduce \textbf{Generation-to-Understanding (G$\rightarrow$U) synergy}.  
Rather than ending reasoning with generation, we reverse the flow.  
Given an input image and a question, the model performs a controlled generative act such as enhancing low-level details, expanding surrounding context, or visualising structural relations.  
The resulting self-generated image, which we call a \textbf{visual thought}, is then fed back into the model as additional evidence for reasoning.  
Generation becomes an internal analytic step that refines perception before the model answers.  
This mechanism is realised entirely through a two-stage zero-shot prompting framework without retraining or external tools.  
The same model generates and understands, forming a closed feedback loop inside the existing architecture.

Extensive experiments across twelve benchmarks show that reversing the flow exposes a strong and previously hidden coupling between synthesis and perception.  
Generative fidelity tightly bounds perceptual gain, and different families of edit prompts exhibit distinct transfer patterns across task types.  
We further test whether models can autonomously decide what to imagine.  
Although they can produce plausible edits, these self-generated visual thoughts lack stable task alignment and often fail to target the information needed for reasoning.  
This reveals a missing component in current unified cognition: models can imagine, but cannot yet reason about how imagination should be used.

Our contributions are threefold:
\begin{itemize}
\item We analyse the intrinsic asymmetry of current unified models and explain why their understanding-to-generation paradigm prevents reciprocal interaction between perception and synthesis.
\item We introduce \textbf{Generation-to-Understanding (G$\rightarrow$U) synergy}, showing how controlled visual generation can serve as an internal reasoning step that enhances perception in a zero-shot, self-contained manner.
\item We establish the first operational framework and systematic study of this reversed paradigm, revealing the governing factors that define when and how imagination improves multimodal understanding.
\end{itemize}

This work reframes multimodal unification from shared architecture to shared cognition and suggests that imagination is not the end of understanding but its beginning.

\section{Related Work}
\subsection{Unified Multimodal Models} 
Recent progress in unified multimodal models has been rapid, revealing different design philosophies for integrating understanding and generation. In terms of design philosophy, these efforts generally follow two distinct paths.
The first path, which adopts a functionally separate design, utilizes an external diffuser. This design connects a pre-trained LLM/VLM backbone to a separate diffusion module, often via lightweight, trainable adapters. The LLM generates compressed latent tokens as ``semantic condition" signals for the external diffuser~\cite{pan2025metaquery,ge2024seed,sun2024generative,tong2025metamorph}. However, we argue this functional separation fundamentally limits deep, bidirectional knowledge interaction.
The second path employs a single transformer architecture to handle both tasks. This path itself has two main branches. One branch is Visual Autoregressive Generation, which leverages discrete visual tokenizers and unifies both text and visual tokens under a single Next-Token-Prediction paradigm~\cite{chen2025januspro,lu2024unified,qu2025tokenflow,team2024chameleon,wang2024emu3,wu2025janus}. The other branch is the Integrated Diffusion, which unifies LLM-based autoregressive reasoning and diffusion-based image synthesis within a single, shared transformer architecture~\cite{liang2025mixture,ma2025janusflow,zhoutransfusion,shi2024lmfusion}.
We posit that this latter design is the most promising foundation for exploring true, reciprocal synergy, as the model natively possesses both reasoning and high-fidelity synthesis faculties. Therefore, we select BAGEL~\cite{deng2025emerging} as our baseline. As a state-of-the-art model built on the integrated transformer solution, it is the ideal platform for our exploration.
\subsection{Thinking with Images} 
Recent advances in multimodal reasoning have garnered widespread attention. One line of work attempts to ``textualize'' this process by reformulating multi-hop visual localization as an explicit, long-form image-text interleaved chain-of-thought~\cite{qicogcom,zheng2025deepeyes,li2025dyfo,su2025pixel,wu2024v,zhang2025chain}. Subsequently, multimodal reasoning is shifting from ``Thinking about Images" to ``Thinking with Images," which transforms vision into a dynamic, manipulable cognitive workspace. One major branch of this paradigm relies on external execution. This includes agentic frameworks that orchestrate a fixed inventory of external visual tools (e.g., OCR, object detectors, or ``chain-of-focus" zoom-ins)~\cite{su2025openthinkimg,hu2024visual,liu2025visual,wu2025vtool}, as well as programmatic approaches where the model generates executable code (e.g., Python scripts) that must be run by an external interpreter~\cite{hong2025deepeyesv2,zhao2025pyvision,qiao2025v,wu2025vtool}.
Our method proposes a distinct paradigm. Instead of outsourcing to text or external tools, we leverage the model's native, intrinsic generative capabilities as the mechanism for ``visual thought". Theoretically, the diversity of these generative functions (e.g., free-form manipulation, enhancement, and editing) allows us to map a much broader spectrum of reasoning processes, enabling not just thinking with images, but true ``imagination" beyond them.
\section{Method}
\label{sec:method}
\begin{figure}[t]
\begin{center}
   \includegraphics[width=0.95\linewidth]{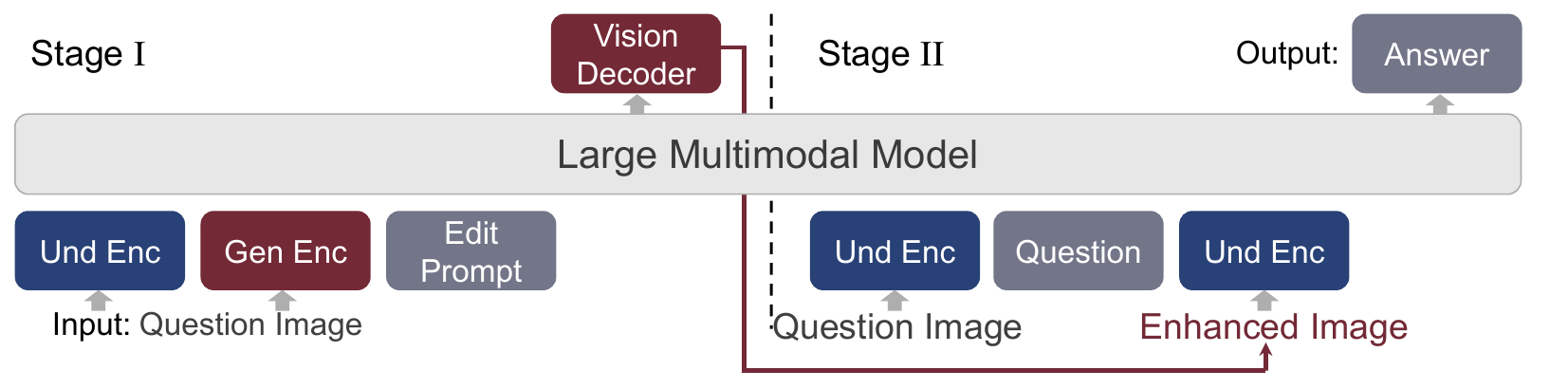}

\caption{Overview of our Generation-to-Understanding (G$\rightarrow$U) framework, a two-stage process. In Stage I (Visual Thought Generation), the unified model leverages its intrinsic generative capabilities to produce an enhanced image. This step functions as an internal analytic process that reconstructs or refines visual evidence. In Stage II (Understanding), this generated ``visual evidence" is concatenated with the original inputs, serving as an explicit visual thinking process to bolster the model's final perception.} \label{fig:method}
\vspace{-3mm}
\end{center}
\vspace{-3mm}
\end{figure}
We view Generation to Understanding (G$\rightarrow$U) not as a prompting heuristic but as a cognitive reformulation of how multimodal models reason.  
In this paradigm, generation is reinterpreted as an internal analytic process that constructs visual evidence before comprehension, rather than a post hoc act of synthesis.  
This section formalises this mechanism and describes how it can be realised within existing large multimodal models without retraining.

\subsection{Concept and Motivation}

Contemporary large multimodal models (LMMs) have achieved remarkable progress in 
\textbf{understanding to generation (U$\rightarrow$G)} tasks such as text guided image editing and synthesis.  
Given an image $I$ and an instruction or prompt $q$, the model first interprets the scene and then generates a modified image $\hat{I}$ according to the described semantics:
\begin{equation}
\hat{I} = \mathcal{D}_G(f_v(I), q),
\end{equation}
where $f_v$ denotes the visual encoder and $\mathcal{D}_G$ the diffusion or autoregressive generator.  
This pipeline represents the dominant one-way paradigm in which understanding drives generation.  
It enables precise visual control, such as adding, removing, or modifying objects, but the generative process itself never contributes back to perception.  
Although these models unify perception and generation at the architectural level, they remain cognitively asymmetric because generation benefits from understanding, while understanding gains nothing from generation.

We reverse this direction and explore the complementary question:  
\textit{can generation itself enhance understanding?}  
We propose the \textbf{Generation to Understanding (G$\rightarrow$U)} framework,  
in which the same model uses its generative capability to imagine auxiliary visual evidence and reuse it to refine perception.  
Generation is therefore redefined as an internal reasoning step, a form of visual thinking that bridges imagination and comprehension.

\subsection{Framework Overview}

Let $\mathcal{M}$ denote a unified multimodal model composed of two coupled pathways,  
an understanding pathway $\mathcal{M}_U$ and a generative pathway $\mathcal{M}_G$, both sharing parameters.  
The proposed G$\rightarrow$U framework converts the conventional one-way U$\rightarrow$G process into a closed cognitive loop.  
Formally, given an image $I$ and a textual query $q$, the model performs
\begin{equation}
\hat{I} = \mathcal{M}_G(I, q; p_{edit}), \qquad
a = \mathcal{M}_U(I, \hat{I}, q),
\label{eq:gu}
\end{equation}
where $p_{edit}$ is a structured visual editing prompt.  
The first step uses $\mathcal{M}_G$ to generate an auxiliary image $\hat{I}$ that represents an imagined or modified view of the scene.  
The second step reintroduces $\hat{I}$ into $\mathcal{M}_U$, allowing the model to reason jointly over the observed and imagined evidence to produce the final answer $a$.  
This self-contained loop realizes the proposed \textbf{G$\rightarrow$U synergy}, showing that generation can serve as a mechanism for constructing internal visual evidence that enhances understanding. The overall process is illustrated in Figure~\ref{fig:method}.

\subsection{Stage I: Visual Thought Generation}

The first phase activates the generative pathway $\mathcal{M}_G = \mathcal{D}_G \circ f_v$ to synthesise an auxiliary image $\hat{I}$, termed a \textbf{visual thought}:
\begin{equation}
\hat{I} = \mathcal{D}_G(f_v(I), q, p_{edit}),
\end{equation}
where $\mathcal{D}_G$ denotes the generative decoder conditioned on both the image features and the instruction.  
The generative prompt $p_{edit}$ specifies how the model should transform $I$ to simulate the missing information that may assist understanding.  
We design a library $\mathcal{P}$ of such prompts, grouped into two complementary families:

\begin{itemize}
    \item \textbf{Enhancement Prompts} $\mathcal{P}_E$: low-level visual refinements such as denoising, deblurring, or exposure correction that improve perceptual fidelity and object clarity.
    \item \textbf{Expansion Prompts} $\mathcal{P}_X$: high-level semantic manipulations such as outpainting, background reconstruction, or viewpoint translation that expand contextual awareness.
\end{itemize}

Each generated image $\hat{I}$ can be interpreted as the model’s internal hypothesis of what the scene could look like if ambiguity or occlusion were resolved.  
This process externalizes the model’s latent world knowledge in a visual form, providing explicit evidence for subsequent reasoning.  
All prompts are designed to avoid direct information leakage or trivial answer hints.

\subsection{Stage II: Understanding via Internal Feedback}

After generation, the auxiliary image $\hat{I}$ is reintroduced into the model to form an augmented multimodal context:
\begin{equation}
\mathcal{C} = \{I, \hat{I}, q\}.
\end{equation}
The understanding pathway then encodes both $I$ and $\hat{I}$ to obtain features $z_v = f_v(I)$ and $\hat{z}_v = f_v(\hat{I})$,  
which are concatenated as $[z_v,q, \hat{z}_v]$ and fed into the reasoning decoder:
\begin{equation}
a = \mathcal{D}_U([z_v,q, \hat{z}_v] ).
\end{equation}
This internal feedback loop $\{I \!\rightarrow\! \hat{I} \!\rightarrow\! a\}$ allows the model to integrate observed and imagined evidence within the same representational space.  
It enhances perception in two complementary ways.  
First, \textit{perceptual enrichment}, where enhancement edits sharpen fine-grained structures and reduce ambiguity.  
Second, \textit{contextual supplementation}, where semantic expansions add missing spatial or relational cues that support high level reasoning.  
When $\hat{I}=I$, the process naturally degenerates to the standard U$\rightarrow$G baseline, ensuring full backward compatibility.

\subsection{Autonomous Prompt Generation}

While hand crafted prompts are effective, they do not scale to diverse reasoning tasks.  
We therefore introduce an \textbf{autonomous prompt writer} $\mathcal{W}$ based on GPT 4o-mini.  
Given $K$ example triplets $\{(I_i, q_i, p_i)\}_{i=1}^{K}$, the writer produces task specific edit prompts through in context learning~\cite{brown2020language}:
\begin{equation}
p_{edit} = \mathcal{W}(I, q;\{(I_i, q_i, p_i)\}_{i=1}^{K}).
\end{equation}
The writer generalises edit semantics to new tasks while preventing answer leakage.  
We use $K{=}5$ demonstrations by default and filter generated prompts by semantic consistency and lexical diversity to ensure robustness.  
This mechanism provides an adaptive way for the model to decide how to imagine, establishing an early step toward self reflective visual reasoning.

\subsection{Implementation and Efficiency}

We instantiate the proposed framework on \textbf{BAGEL (7B)}.  
Its integrated Transformer backbone couples diffusion based synthesis and autoregressive reasoning within shared self attention layers.  
This architecture naturally supports bidirectional information flow between $\mathcal{M}_G$ and $\mathcal{M}_U$,  
making it an ideal testbed for the G$\rightarrow$U formulation.  
All experiments are conducted in a zero shot setting without any fine tuning or additional parameters.   
This lightweight design demonstrates that imagination driven feedback can be realised entirely within existing unified models without retraining or architectural modification.

\section{Experiments}

The central question of this work is simple yet fundamental: \textit{can generation think?}  
If large multimodal models can already understand to generate, can they also generate to understand?  
We test this hypothesis by examining whether controlled visual generation, which we term \textit{visual thinking}, can act as an internal reasoning process that enhances perception and cognition.  
Our investigation proceeds in two complementary phases.  
First, we conduct an exploratory study to verify the existence of Generation to Understanding (G$\rightarrow$U) synergy and identify which forms of generative edits most effectively enhance comprehension.  
Second, we perform a large scale quantitative analysis across twelve benchmarks to measure how the form and fidelity of generation translate into measurable understanding gains.
\begin{table}[h!]
\caption{Statistics of our curated VisThink-Bench. The benchmark contains $1595$ samples, sourced from $12$ existing benchmarks and organized into $3$ main categories (Perceptual, Logical Reasoning, and Spatial Reasoning) and $34$ fine-grained sub-tasks.}
\scriptsize
\label{tab:task_summary}
\centering
\scalebox{0.95}{ 
\renewcommand{\tabularxcolumn}[1]{m{#1}} 
\begin{tabularx}{\columnwidth}{ 
    >{\raggedright\arraybackslash}m{2cm} | 
    >{\raggedright\arraybackslash}X |    
    c                                
}
\toprule
\textbf{Tasks} & \textbf{Source} & \textbf{Total} \\
\midrule
Perception & MME~\cite{fu2023mme}, MMBench~\cite{liu2024mmbench}, SEED~\cite{li2024seed}, MMStar~\cite{chen2024mmstar}, R-Bench~\cite{rbench}, Q-Bench~\cite{wu2023qbench} & $970$ \\
\midrule 
Logic Reasoning & EMMA~\cite{hao2025can},  MMBench~\cite{liu2024mmbench}, SEED~\cite{li2024seed}, KiVA~\cite{yiukiva}, MMStar~\cite{chen2024mmstar}, HallusionBench~\cite{guan2024hallusionbench} & $317$ \\
\midrule
Spatial Reasoning & LogicVista~\cite{xiao2024logicvista}, LEGO~\cite{tang2025lego}, Spatial-457~\cite{wang2025spatial457}, MMBench~\cite{liu2024mmbench} & $308$ \\
\bottomrule
\end{tabularx}
} 
\end{table}

\begin{figure*}[t]
\begin{center}
\includegraphics[width=0.95\linewidth]{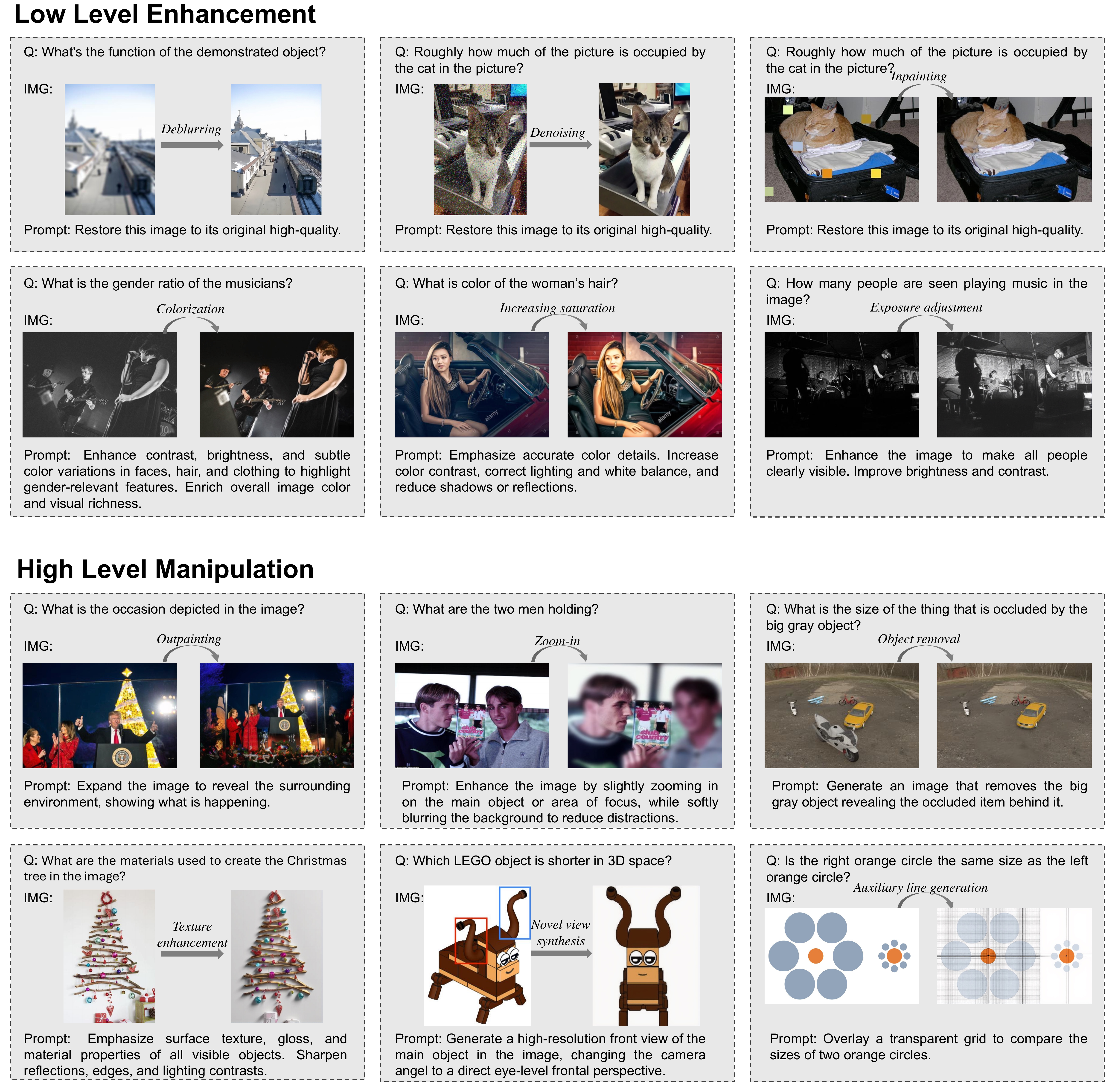}

\vspace{-1mm}
\caption{Visualization of the two functional regimes of visual editing that enable our G$\rightarrow$U framework. Top: \textbf{Low-level enhancements} (e.g., \textit{deblurring, denoising, inpainting, colorization, increasing saturation, exposure adjustment}) utilize the model's generative prior to refine raw visual quality and enhance perceptual signals. Bottom: \textbf{High-level manipulations} (e.g., \textit{outpainting, zoom-in, object removal, texture enhancement, novel view synthesis, auxiliary line generation}) leverage the model's internal world model to expand semantic context and simulate counterfactuals as an intermediate reasoning step. All depicted examples showcase initial BAGEL \textit{failures} that our G$\rightarrow$U visual thinking successfully corrects.
} \label{fig:edit_vis}
\end{center}
\vspace{-8mm}
\end{figure*}
\begin{figure*}[t]
\begin{center}
\includegraphics[width=1.0\linewidth]{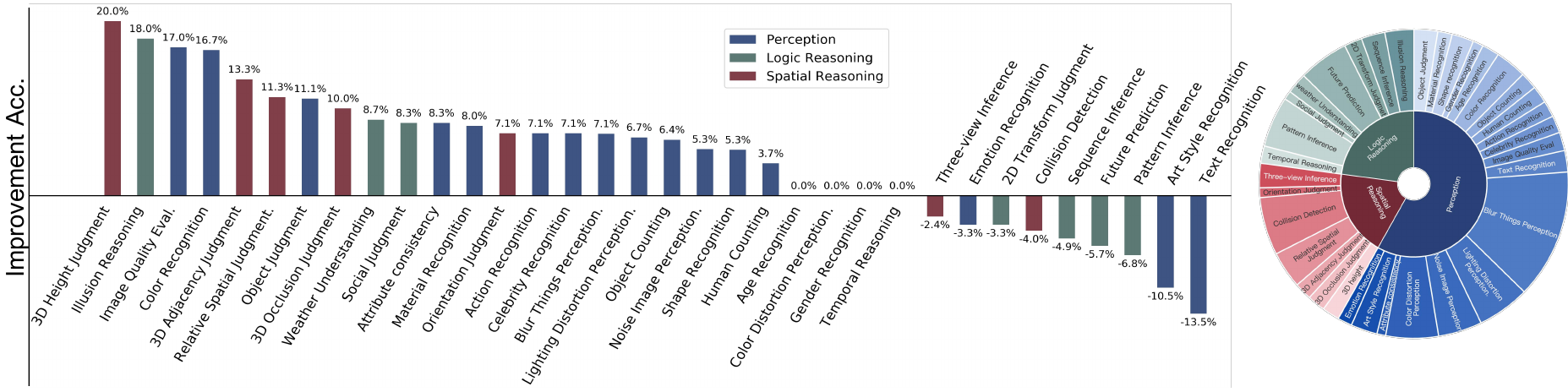}

\caption{Quantitative evaluations on our VisThink-Bench. Left: We report the accuracy change of our G$\rightarrow$U framework relative to the vanilla BAGEL baseline across all fine-grained tasks. Right: The task composition of VisThink-Bench, which spans $3$ primary categories (Perceptual, Logical Reasoning, and Spatial Reasoning) and $34$ fine-grained sub-tasks.}
 \label{fig:our_result}
\end{center}
\vspace{-8mm}
\end{figure*}
\subsection{Setting}

\noindent \textbf{Evaluation Data.}  
To explicitly examine how visual generation contributes to understanding, we constructed a dedicated evaluation suite named \textbf{VisThink-Bench}.  
It contains $1595$ carefully curated VQA samples drawn from $34$ tasks across $12$ established benchmarks.  
Each task was selected for its fine grained correspondence between visual evidence and reasoning demand, enabling a direct mapping from specific generative edits to their impact on comprehension.  
The tasks cover both foundational perception such as colour or attribute recognition and higher level reasoning such as spatial or causal inference.  
The overall task composition and data sources are summarised in Table~\ref{tab:task_summary} and Figure~\ref{fig:our_result} (right).

In addition to VisThink-Bench, we evaluate our framework on seven widely used benchmarks across five capability axes.  
For general perception and composition, we use MMBench~\cite{liu2024mmbench}, MME~\cite{fu2023mme}, and MM-Vet~\cite{yu2023mmvet}.  
For reasoning, we adopt MM-Star~\cite{chen2024mmstar} and KiVA~\cite{yiukiva}.  
For hallucination mitigation, we use HallusionBench~\cite{guan2024hallusionbench}.  
For robustness, we employ R-Bench~\cite{rbench}, which measures model stability under various visual corruptions.  
Together these datasets form a comprehensive test bed to evaluate how G$\rightarrow$U synergy generalises from perception to cognition.

\noindent \textbf{Implementation Details.}  
For VisThink-Bench, we manually designed task specific edit prompts that align each question type with its corresponding visual operation.  
For other benchmarks, prompts were generated automatically using a few shot writer based on GPT 4o-mini with five in context examples.  
All visual edits were executed using BAGEL’s default diffusion settings with $30$ denoising steps and classifier free guidance scales of $4.0$ for text and $1.0$ for image.  

\noindent \textbf{Comparison Methods.}  
We compare the proposed G$\rightarrow$U framework with both the vanilla BAGEL~\cite{deng2025emerging} baseline and a range of state of the art multimodal models.  
For specialised understanding systems, we include Qwen2.5 VL~\citep{qwen25vl}, InternVL2.5~\citep{internvl25}, LLaVA OV~\citep{llava}, and DeepSeek VL2~\citep{dsvl2}.  
For unified models, we cover diffusion based architectures such as MetaQuery~\citep{pan2025metaquery} and BLIP3-o~\citep{chen2025blip3o}, as well as hybrid autoregressive and diffusion models such as Janus Pro, TokenFlow XL~\citep{qu2025tokenflow}, Show o2~\citep{xie2025showo2}, and LlamaFusion~\citep{shi2024lmfusion}.  
This selection ensures comprehensive coverage of current multimodal paradigms across both understanding and generation domains.

\subsection{Editing Task Scope}

To trace how generation contributes to perception, we begin with an exploratory analysis of visual editing behaviours.  
As illustrated in Figure~\ref{fig:edit_vis}, effective edits cluster into two functional regimes with complementary roles in the G$\rightarrow$U process.

\noindent \textbf{Low level enhancement.}  
These operations utilise the model’s intrinsic generative prior to refine raw visual quality through deblurring, denoising, or exposure adjustment.  
They enhance perceptual signals before reasoning, sharpening contours and restoring contrast to reduce ambiguity.  
Such refinement directly improves tasks that rely on local visual evidence such as object counting, attribute recognition, or colour reasoning, and provides a consistent foundation for understanding within the G$\rightarrow$U loop.

\noindent  \textbf{High level manipulation.}  
Beyond restoration, the model performs structured edits that reflect its internal world model, such as outpainting, removing distractors, zoom in, texture enhancement, novel view synthesis, or auxiliary line generation (Figure~\ref{fig:edit_vis}, bottom).  
These manipulations expand semantic context and simulate counterfactuals, enabling the model to hypothesise and verify possible interpretations.  
By transforming imagination into an intermediate reasoning step, high level manipulation bridges generative perception with analytical understanding.

\subsection{Results and Analysis}

\noindent \textbf{VisThink-Bench.}  
Figure~\ref{fig:our_result} shows the relative accuracy gain of G$\rightarrow$U over the vanilla BAGEL baseline across $34$ task categories. 
The pattern reveals a strong alignment between the type of generation and the nature of understanding it benefits.

Perception, Logic, and Spatial reasoning driven tasks such as 3D height estimation, illusion reasoning, and color recognition achieve the largest improvements, exceeding ten percent. These categories depend heavily on local contrast (quality enhancement), spatial layout (novel view), and fine scale details (zoom in), which are precisely the cues strengthened through visual thoughts.
Moderate gains appear in complex scenario perception tasks including human counting, object counting, shape recognition, where complex scenario pose accurate editing a big challenge.
Performance declines slightly in symbol intensive tasks such as text or pattern recognition, where the generative prior lacks discrete token fidelity.  
Overall, these results confirm that controlled generation yields consistent and interpretable improvements in visually grounded reasoning, while revealing its current limits in symbolic abstraction.

\newlength{\mycolumnwidth}
\setlength{\mycolumnwidth}{(\textwidth - 18em) / 8} 
\newcolumntype{E}{>{\centering\arraybackslash}p{\mycolumnwidth}}
\begin{table*}[]
\centering
    \setlength{\tabcolsep}{3pt}
    \renewcommand{\arraystretch}{1.2}
    \scriptsize
    \caption{Quantitative evaluations on MMBench~\citep{liu2024mmbench}, MME~\citep{fu2023mme} (where MME-P denotes the Perception score and MME-S denotes the total Sum score, i.e., Perception + Cognition),  MM-Vet~\citep{yu2023mmvet}, MMStar~\citep{chen2024mmstar}, KiVA~\cite{yiukiva}, HallusionBench~\citep{guan2024hallusionbench} and R-Bench~\citep{rbench}.
    }
    \vspace{-2mm}
    \label{table1}
    \scalebox{0.75}{
\begin{tabular}{p{9em}c*{4}{E}|*{2}{E}|E|E}
\toprule
 &  & \multicolumn{4}{c}{\textit{\textbf{General VQA}}}       & \multicolumn{2}{c}{\textit{\textbf{Logic Reasoning}}} & \multicolumn{1}{c}{\textit{\textbf{Hallucination}}} & \multicolumn{1}{c}{\textit{\textbf{Robustness}}} \\
\rowcolor{gray!20} 
\multicolumn{1}{c}{\cellcolor{white}\multirow{-2}{*}{\textbf{Models}}} & \cellcolor{white}\multirow{-2}{*}{\textbf{\# Params}} & \textbf{MMB} & \textbf{MME-P} & \textbf{MME-S} & \textbf{MMVet} & \textbf{MMStar} & \textbf{KiVA} & \textbf{HallBench}& \textbf{R-Bench} \\
\midrule

        \multicolumn{2}{l}{\textit{Understanding Only}} &&&&&&&&\\
                                
Qwen2.5-VL~\cite{qwen25vl}                                  & $3B$                                                                       
& $79.1$  &  -  &  $2157$  & $61.8$ & $55.9$  &  -   & $46.3$ &  -    \\
InternVL2.5~\cite{internvl25}                                 & $4B$                                                                       
& $81.1$ &   -  & $2337$  & $60.6$& $58.3$ &  -  & $46.3$ & $66.1$    \\
Llava-OV~\cite{llava}                                    & $7B$                                                                       
& $80.8$& $1580$ & $1998$  & $57.5$  &  $61.7$  &  -  &   -  & -   \\
Qwen2.5-VL~\cite{qwen25vl}                                  & $7B $                                                                      
&  $83.5$ &  -  & $ 2347$ &   $67.1$  & $ 63.9$ & -   &  -   &  -   \\
InternVL2.5~\cite{internvl25}                                 & $7B $                                                                      
& $83.6 $ &   -  & $2344 $ & $ 62.8$&   -  &    -   & $52.9$ & -    \\
Emu3-chat~\cite{wang2024emu3}                                   &$ 8B$                                                                       
& $58.5$ &  $1244 $ &  -   & $37.2$  &  - &  -  &  -  &  -    \\

DeepSeek-VL2~\cite{dsvl2}                                & $4.1B/28B $                                                                
& $79.6 $ &  -  &  $2253 $ & $ 60.0$ &  $61.3$  &   -   &  -   & -  \\ 
\midrule
        \multicolumn{2}{l}{\textit{Unified model}} &&&&&&&&\\
MetaQuery-L~\cite{pan2025metaquery}                                 &$ 3B  $                                                                     &       $ 78.6$       & $  1574 $   &    -   &  $63.2$     &     -      &         -      &     -    &          -              \\
BLIP3-o~\cite{chen2025blip3o}                                     &$ 4B $                                                                      &     $  78.6   $     &  $  1527 $  &  $ 2160  $  &   $ 60.1 $  &    -       &       -        &      -   &         -              \\
Janus-Pro~\cite{chen2025januspro}                                   &$ 7B $                                                                      &     $ 79.2$         &  $ 1567 $   &   -    &   $ 50.0 $  &     -      &            -   &    -     &       -                \\
MetaQuery-XL~\cite{pan2025metaquery}                                & $7B $                                                                      &      $   83.5  $    &   $ 1685 $  &    -   &   $66.6 $   &     -      &           -    &     -    &         -           \\
Show-o2~\cite{xie2025showo2}                                     &$ 7B $                                                                      &      $ 79.3  $      &   $1620$    &    -   &     -  &    $  56.6  $   &         -      &       -  &             -          \\
LMFusion~\cite{shi2024lmfusion}                                    &$ 8B $                                                                      &        $72.1$       &   $ 1630  $ &    -   &    -   &      -     &            -   &     -    &            -            \\
TokenFlow-XL~\cite{qu2025tokenflow}                                &$ 13B   $                                                                   &      $ 68.9   $     &  $ 1545  $  &  $  1840  $ & $ 40.7 $    &     -      &        -       &    -     &         -            \\
 \cdashline{1-10}
BAGEL~\cite{deng2025emerging}                                       &$ 7B $                                                                      &    $  83.7 $        &   $ 1686  $ &   $ 2320  $ &    $62.7$   &       $66.7 $   &      $  32.9 $      &  $ 50.9  $    &       $ 70.1  $                                 \\  
BAGEL+Ours                                     &$ 7B  $                                                                     &     $ 85.5  $       &$  1662 $    &   $ 2315 $  & $  62.1$    &     $  67.9 $   &       $35.2 $       &     $ 55.1 $  &        $  71.7  $                             \\  
\bottomrule
\end{tabular}
}
\vspace{-4mm}
\end{table*}

\noindent \textbf{Other Benchmarks.}  
Table~\ref{table1} presents the quantitative comparison across seven standard benchmarks.  
Among unified models, our G$\rightarrow$U enhanced BAGEL achieves gains on a majority of the benchmarks, particularly on reasoning and robustness oriented datasets.  
Performance increases by $1.2\%$ on MMStar, $4.2\%$ on HallusionBench, $1.8\%$ on MMBench, and $1.6\%$ on R-Bench, showing that generative feedback strengthens contextual inference and stabilises perception under distortion.  
While specialised understanding models such as Qwen2.5 VL~\citep{qwen25vl} and InternVL2.5~\citep{internvl25} maintain higher absolute scores, they depend on domain specific tuning and lack cognitive coupling between generation and reasoning.  
Our framework closes this gap without fine tuning, showing that imagination driven feedback offers a general, parameter free route toward more interpretable and resilient multimodal understanding.

\begin{figure}[tbp]
    \centering
    \begin{subfigure}[b]{0.23\textwidth}
        \centering
        \includegraphics[width=\textwidth]{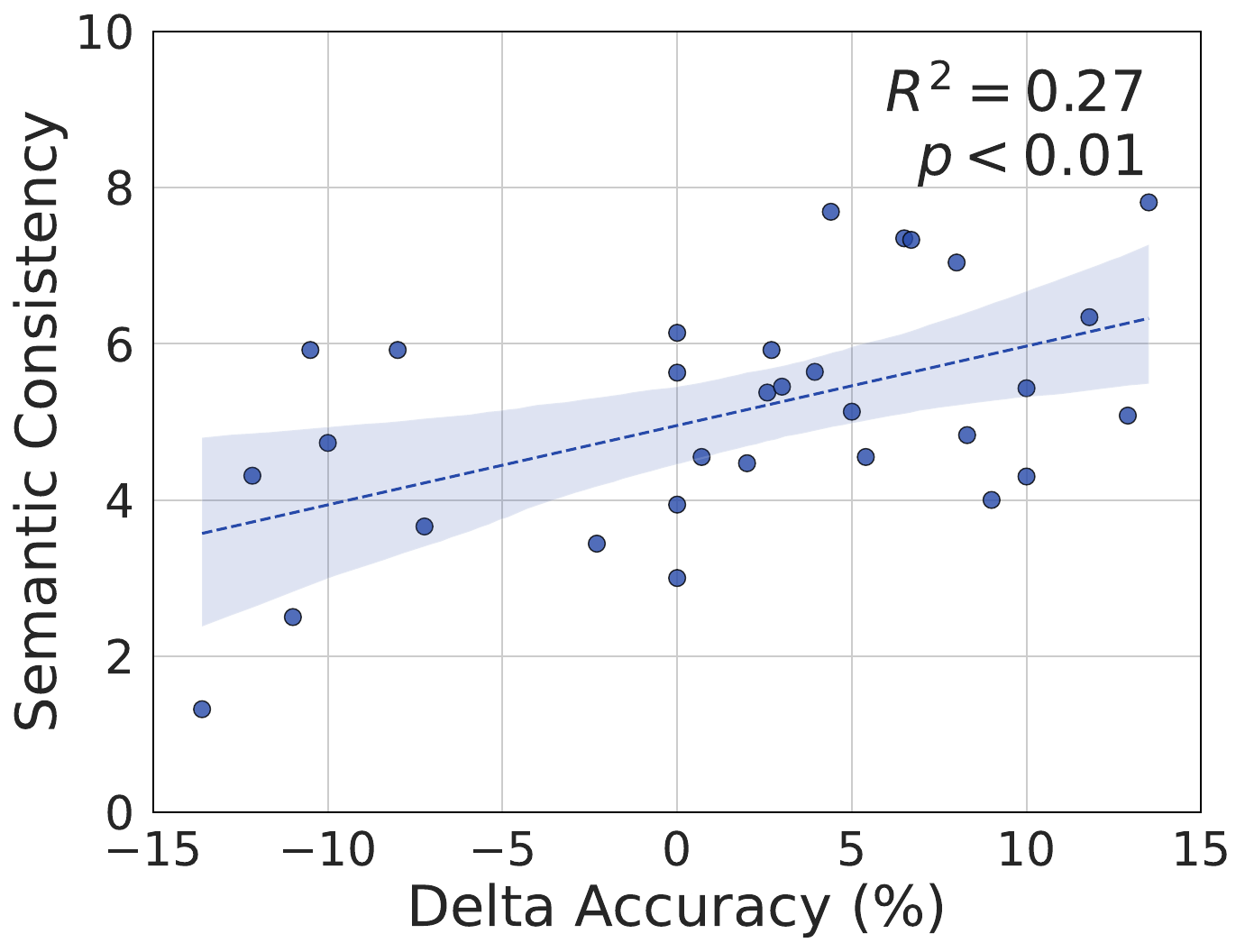} 
    \end{subfigure}
    \begin{subfigure}[b]{0.23\textwidth}
        \centering
        \includegraphics[width=\textwidth]{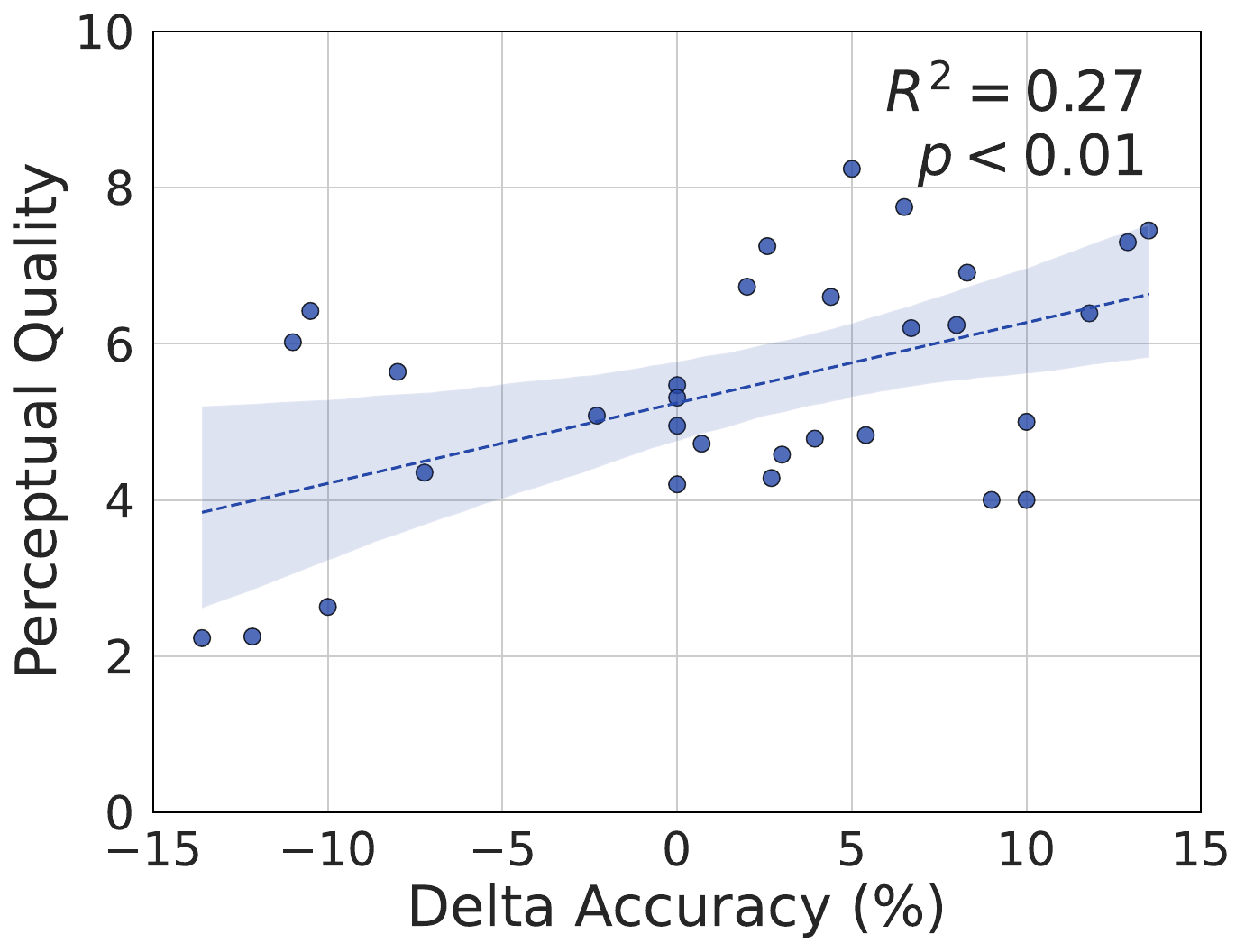} 
    \end{subfigure}
    \vspace{-3mm}
    \caption{Correlation between improvements and editing performance. Linear regression analysis reveals a statistically significant positive correlation between the two variables ($R^2 = 0.27, p < 0.01$). The dashed line represents the linear regression fit, and the shaded area indicates the $95\%$ confidence interval.}  
    \label{g2u_fig_corr}
    \vspace{-3mm}
\end{figure}

\noindent\textbf{Editing Quality.}  
How good do visual thoughts need to be in order to help understanding?  
To find out, we evaluated BAGEL’s editing fidelity on VisThink-Bench using the VIE metric~\cite{ku2024viescore}, obtaining an average Semantic Consistency of $5.12$ and a Perceptual Quality of $5.41$.
We then measured editing quality for each task in VisThink-Bench and correlated it with the accuracy gain produced by G$\rightarrow$U.  
The trend is clear.  
As shown in Figure~\ref{g2u_fig_corr}, both semantic consistency and perceptual quality exhibit a statistically significant positive correlation with downstream improvement ($R^2 = 0.27$, $p < 0.01$).  
In other words, better visual thoughts tend to deliver better understanding.
The correlation is moderate, which is exactly what we expect: generation helps, but the size of the benefit still depends on the task and the prompt.  
Even so, the pattern reinforces the intuition behind G$\rightarrow$U — when the model imagines well, it understands better.

\vspace{-1mm}
\section{Further Analysis}
To uncover the mechanism behind Generation to Understanding (G$\rightarrow$U) synergy, we analyse how architectural design, reasoning modality, and prompting behaviour shape its effectiveness.  
The goal is not only to test robustness but to reveal what enables generation to function as reasoning.

\subsection{Architectural Robustness}
\noindent \textbf{Visual versus textual reasoning.}  \label{sec:textcot}
We first compare our visual thinking process with BAGEL’s textual chain-of-thought (CoT) reasoning.  
As shown in Table~\ref{tab:more_ab}, textual CoT reduces average accuracy from $62.6$ to $57.8$, confirming that verbal reasoning introduces spurious linguistic bias when repurposed for visual comprehension.  
Visual thinking instead performs reasoning in the image space, transforming generation itself into an internal analytic step.  
It is faster, more stable, and better aligned with perception.  
This result isolates a key distinction: textual CoT explains decisions post-hoc, whereas visual thinking performs reasoning pre-hoc by reshaping perception before understanding.

\noindent \textbf{Context integration.}  
We next examine how the generated visual clue should be injected into the model’s context.  
Variant \textcircled{1} Replace substitutes the original image with the edited one.  
Variant \textcircled{2} Concat places both images sequentially.  
Variant \textcircled{3} VAE Concat merges generative and semantic features at the representation level.  
Table~\ref{tab:more_ab} shows that both Replace and Concat outperform the baseline, confirming the robustness of the G$\rightarrow$U mechanism.  
Concat yields the highest gain ($63.5$ average), while VAE Concat collapses ($59.1$) due to instruction failure.  
After using VAE features, the model suffers from modality confusion, failing to understand and answer questions because it struggles to determine whether it should generate or comprehend. This is a severe modal confusion problem.

\subsection{Prompt Writer Behaviour}

\noindent\textbf{Can the model decide what to imagine?}  
A central question for G$\rightarrow$U is whether a unified model can autonomously choose the type of visual edit that will improve its own understanding.  
To assess this, we compare three prompt writers:  
(1) the model itself, referred to as Self-Prompt,  
(2) Gemini 2.5 Flash,  
and (3) our few-shot GPT 4o mini writer.

\begin{figure}[t]
\begin{center}
   \includegraphics[width=0.95\linewidth]{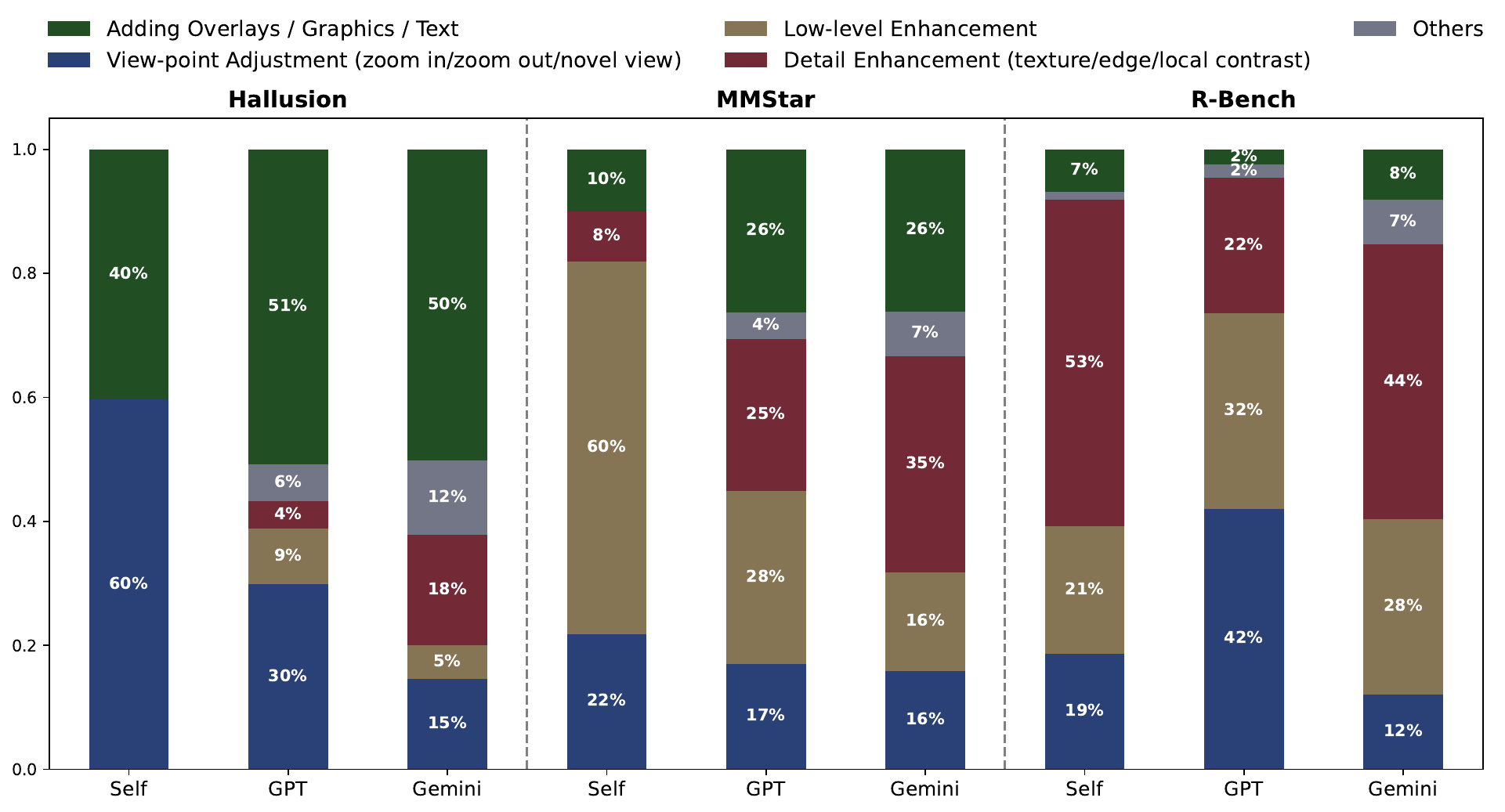}
\vspace{-2mm}
\caption{Comparison of generated edit prompt behaviors} \label{fig:prompt}
\vspace{-8.1mm}
\end{center}
\end{figure}

\noindent\textbf{Self-Prompt shows partial ability but lacks task awareness.}  
As shown in Table~\ref{tab:more_ab}, Self-Prompt achieves reasonable performance with an average score of $63.4$.  
This indicates that the model can construct syntactically valid and visually coherent edit instructions.  
However, its prompts often focus on superficial changes rather than the information that would genuinely help the task. As shown in Figure~\ref{fig:prompt}, Self-Prompt's behavior is less diverse and fails to adapt strategically to the task requirements. 
This results in inconsistent improvements across benchmarks and reveals that, although the model can generate plausible edits, it does not know which edits would actually enhance understanding.

\noindent\textbf{External writers provide more reliable and task-aligned guidance.}  
Gemini 2.5 Flash produces competitive results and improves performance over the baseline in most settings.  
GPT 4o mini performs best overall with an average score of $64.9$, showing strong alignment between its edit instructions and the reasoning demands of the tasks.  
Its prompts consistently yield visual thoughts that supply useful complementary evidence for the model.

\noindent\textbf{Implication: imagination is available but not directed.}  
These findings show that unified models possess the generative skills required for visual thinking, yet they lack the meta-awareness needed to guide those skills.  
They can imagine when instructed, but they cannot reliably determine what to imagine.  
This gap marks a key limitation in current multimodal intelligence: the model can create new visual evidence, but it cannot yet reason about how imagination should be used to support understanding.

\begin{table}[t]
\caption{Ablation studies, investigating the impact of different context concatenation strategies for integrating the ``visual thought" and the influence of various prompt writers on overall performance.}\vspace{-2mm}
    \scriptsize
    \label{tab:more_ab}
     \centering
    \scalebox{0.85}{
    \begin{tabular}{l cc  cc }
    \toprule
    Method  & \textbf{R-Bench}~\citep{rbench} & \textbf{HallBench}~\citep{guan2024hallusionbench} & \textbf{MMStar}~\citep{chen2024mmstar} & \textbf{AVG}\\
    \midrule
    BAGEL (Baseline) & $70.1$ & $50.9$ &  $66.7$ & $62.6$ \\
    BAGEL Textual CoT & $63.6$ & $50.4$ & $59.4$ & $57.8$\\
    \rowcolor{gray!20} Architectural Ablation & &  && \\
   \textcircled{1} Replace & $70.1$ & $50.5$ &  $67.2$ & $62.6$ \\
   \textcircled{2} Concat & $70.9$ & $53.1$ & $66.5$ & $63.5$  \\
   \textcircled{3} VAE Concat & $69.9$ & $42.2$ & $65.2$ & $59.1$  \\
   \rowcolor{gray!20} Prompt Writer Ablation & &  && \\
   \textcircled{4} Self-Prompt & $70.1$ & $53.3$  & $66.8$ & $63.4$ \\
   \textcircled{5} Gemini-2.5-Flash & $72.9$ & $52.4$ & $67.7$ & $64.3$\\
   \textcircled{6} GPT-4o-mini (Ours) & $71.7$ & $55.1$ & $67.9$ & $64.9$\\
   
    \bottomrule
    \end{tabular}
    }\vspace{-0.5em}
    \vspace{-0.5em}
    
\end{table}

\section{Limitations and Failure Modes}

Although Generation to Understanding (G$\rightarrow$U) synergy provides clear benefits, our analysis reveals three characteristic failure modes that define its boundaries and expose the limits of current unified cognition.

\noindent \textbf{1. Generative fidelity as an upper bound.}  
The effectiveness of G$\rightarrow$U is constrained by the model’s intrinsic generative fidelity.  
When the generator fails to reproduce fine-grained or symbolic details such as text, charts, or small numeric patterns, the visual thought adds no new evidence for reasoning.  
In these cases, generation becomes repetition rather than reflection, indicating that understanding cannot exceed what imagination can faithfully reconstruct.

\noindent \textbf{2. Circular reasoning in abstract prompts.}  
Ambiguous or high-level prompts such as “extract the most salient object” or “generate a painting in the same style” often lead to failure.  
Such instructions require prior semantic understanding to perform the generation itself, creating a self-referential loop.  
The model cannot generate what it does not comprehend, so the generative act provides no additional insight.  
This result highlights a critical asymmetry: visual thinking succeeds only when generation introduces new perceptual evidence rather than mirroring existing bias.

\noindent \textbf{3. Lack of extrapolative imagination.}  
Current models exhibit interpolative rather than predictive reasoning.  
Prompts that demand causal anticipation or temporal simulation, such as forecasting motion or inferring future events, consistently fail.  
The model can rearrange what it has observed but cannot project what it has not, revealing the absence of causal world models within today’s multimodal systems.  
This limitation defines the point where visual thinking ceases to be reasoning and reverts to pattern completion.

Together these observations delineate the cognitive frontier of current multimodal intelligence: imagination can inform perception, but only within the limits of fidelity, grounding, and causality.

\section{Conclusion}
This work asked a simple question: can generation think.  
We introduced Generation to Understanding (G$\rightarrow$U) synergy, where visual generation becomes an internal reasoning step that refines perception.  
Experiments across twelve benchmarks confirm that imagination-driven feedback improves multimodal understanding without extra training.  
Our analysis further shows that this synergy is robust yet bounded by fidelity, grounding, and causality.  
The findings suggest that imagination is not the end of understanding but its beginning, and that the next step toward unified cognition is the ability to decide what to imagine.

\section{Acknowledgment}
This work was supported by the National Natural Science Foundation of China (Grant 62406171, 62225601, U23B2052,62306031), in part by the Beijing Natural Science Foundation Project No. L242025, in part by the  Guizhou Province Science and Technology Plan Project (No. QianKeHeZhongDa [2025]031) ,  and in part by the Fundamental Research Funds for the Beijing University of Posts and Telecommunications under Grant  2025AI4S15, in part by the Beijing Key  Laboratory  of Multimodal Data  Intelligent Perception  and Governance,  in part by BUPT Excellent Ph.D. Students Foundation No. CX2023112. The authors thank Ruoyi Du for
his valuable feedback and discussion.

{
    \small
    \bibliographystyle{ieeenat_fullname}
    \bibliography{main}
}

\end{document}